\documentclass[runningheads]{llncs}

\usepackage[T1]{fontenc}
\usepackage{amsmath,amssymb,amsfonts}
\usepackage{graphicx}
\usepackage{textcomp}
\usepackage{marvosym}
\usepackage{xcolor}
\usepackage{booktabs}
\usepackage{multirow}
\usepackage{pifont}
\usepackage{placeins}
\usepackage{float}
\usepackage[hidelinks]{hyperref}
\usepackage{enumitem}
\setlist[itemize]{label=\textbullet,leftmargin=1.5em,itemsep=1.5pt,parsep=0pt,topsep=3pt,partopsep=0pt}
\AtBeginDocument{%
  \setlength{\abovedisplayskip}{6pt plus 2pt minus 2pt}%
  \setlength{\belowdisplayskip}{6pt plus 2pt minus 2pt}%
  \setlength{\abovedisplayshortskip}{4pt plus 2pt minus 2pt}%
  \setlength{\belowdisplayshortskip}{4pt plus 2pt minus 2pt}%
}

\newcommand{\xmark}{\ding{55}}
\newcommand{\cmark}{\ding{51}}
\providecommand{\best}[1]{\textbf{#1}}
\providecommand{\second}[1]{\underline{#1}}

\begin{document}

\title{UHR-Net: An Uncertainty-Aware Hypergraph Refinement Network for Medical Image Segmentation}
\titlerunning{UHR-Net for Medical Image Segmentation}

\author{Shuokun Cheng \and Jinghao Shi \and Kun Sun\textsuperscript{(\Letter)}}
\authorrunning{S. Cheng et al.}

\institute{School of Computer Sciences, China University of Geosciences (Wuhan),\
Wuhan 430074, Hubei, China\\
\email{\{chenyuekun, shijinghao, sunkun\}@cug.edu.cn}}

\maketitle

\begin{abstract}
Accurate lesion segmentation is crucial for clinical diagnosis and treatment planning. However, lesions often resemble surrounding tissues and exhibit ill-defined boundaries, leading to unstable predictions in boundary/transition regions. Moreover, small-lesion cues can be diluted by multi-scale feature extraction, causing under- or over-segmentation. To address these challenges, we propose an Uncertainty-Aware Hypergraph Refinement Network (UHR-Net). First, we introduce an Uncertainty-Oriented Instance Contrastive (UO-IC) pretraining strategy that couples geometry-aware copy-paste augmentation with hard-negative mining of lesion-like background regions to improve instance-level discrimination for small and visually ambiguous lesions. Second, we design an Uncertainty-Guided Hypergraph Refinement (UGHR) block, which derives an entropy-based uncertainty map from a coarse probability map to guide hypergraph refinement. By splitting hyperedge prototypes into foreground and background groups, UGHR decouples higher-order interactions and improves refinement in ambiguous regions. Experiments on five public benchmarks demonstrate consistent gains over strong baselines. Code is available at: \url{https://github.com/CUGfreshman/UHR-Net}.

\end{abstract}

\section{Introduction}
\label{sec:intro}
Medical image segmentation provides pixel-level anatomical or pathological delineation for clinical image interpretation. For lesion analysis, accurately separating lesion areas from surrounding tissues is especially important, because subsequent measurements, visual assessment, and computer-aided decisions all depend on reliable boundary localization.

Although deep neural networks have greatly improved medical segmentation performance, lesion segmentation in realistic scenarios is still challenging. One difficulty comes from scale imbalance: small lesions contribute only limited foreground evidence and may be weakened when features are repeatedly aggregated across scales. Another difficulty lies in local ambiguity. Lesion boundaries often change gradually and may share similar appearance with nearby tissues, so the model can confuse uncertain boundary pixels with either foreground or background regions.

Prior work has attempted to improve lesion segmentation from several directions. Convolutional networks mainly enhance local details and fuse multi-scale features~\cite{yu2023hardnet,zhang2022lesion}, whereas Transformer- and graph-based models introduce broader contextual interactions through attention or message passing~\cite{he2023h2former,jiang2023vig}. Other studies use uncertainty estimates to locate unreliable regions or adjust the training objective~\cite{kendall2017uncertainties,tang2022unified}. Nevertheless, uncertainty is often treated as an auxiliary confidence cue, rather than being directly embedded into a structured reasoning process. Meanwhile, hypergraphs are suitable for modeling relations among multiple nodes, but their use in medical segmentation has rarely been coupled with uncertainty-aware foreground/background reasoning~\cite{gao2022hgnnplus}.

In this paper, we propose UHR-Net, an Uncertainty-Aware Hypergraph Refinement Network for lesion segmentation. To alleviate small-lesion cue dilution and lesion-like background interference, we introduce Uncertainty-Oriented Instance Contrastive (UO-IC) pretraining, which constructs positive lesion pairs through geometry-aware copy-paste augmentation and mines lesion-like backgrounds as hard negatives. To improve prediction consistency in ambiguous regions, we further design an Uncertainty-Guided Hypergraph Refinement (UGHR) block, which derives an entropy-based uncertainty map from a coarse probability map to modulate node-hyperedge participation. Foreground- and background-conditioned hyperedge prototypes are used to decouple higher-order aggregation pathways, reducing boundary interference and strengthening refinement in uncertain regions. Extensive experiments on ISIC-2016, ISIC-2017, GlaS, Kvasir-SEG, and Kvasir-Sessile demonstrate consistent gains over representative baselines.

The main contributions of this paper are summarized as follows:
\begin{itemize}
\item We propose UHR-Net, an uncertainty-aware hypergraph refinement framework for lesion segmentation that integrates contrastive pretraining with structured higher-order refinement.
\item We introduce UO-IC pretraining, which combines geometry-aware copy-paste augmentation and lesion-like hard-negative mining to enhance instance-level discrimination for small and ambiguous lesions.
\item We develop the UGHR block, which uses entropy-derived uncertainty to guide hypergraph refinement and employs foreground/background-conditioned hyperedge prototypes to decouple higher-order interactions in ambiguous regions.
\end{itemize}

\begin{figure}[t]
  \centering
  \includegraphics[width=0.96\textwidth]{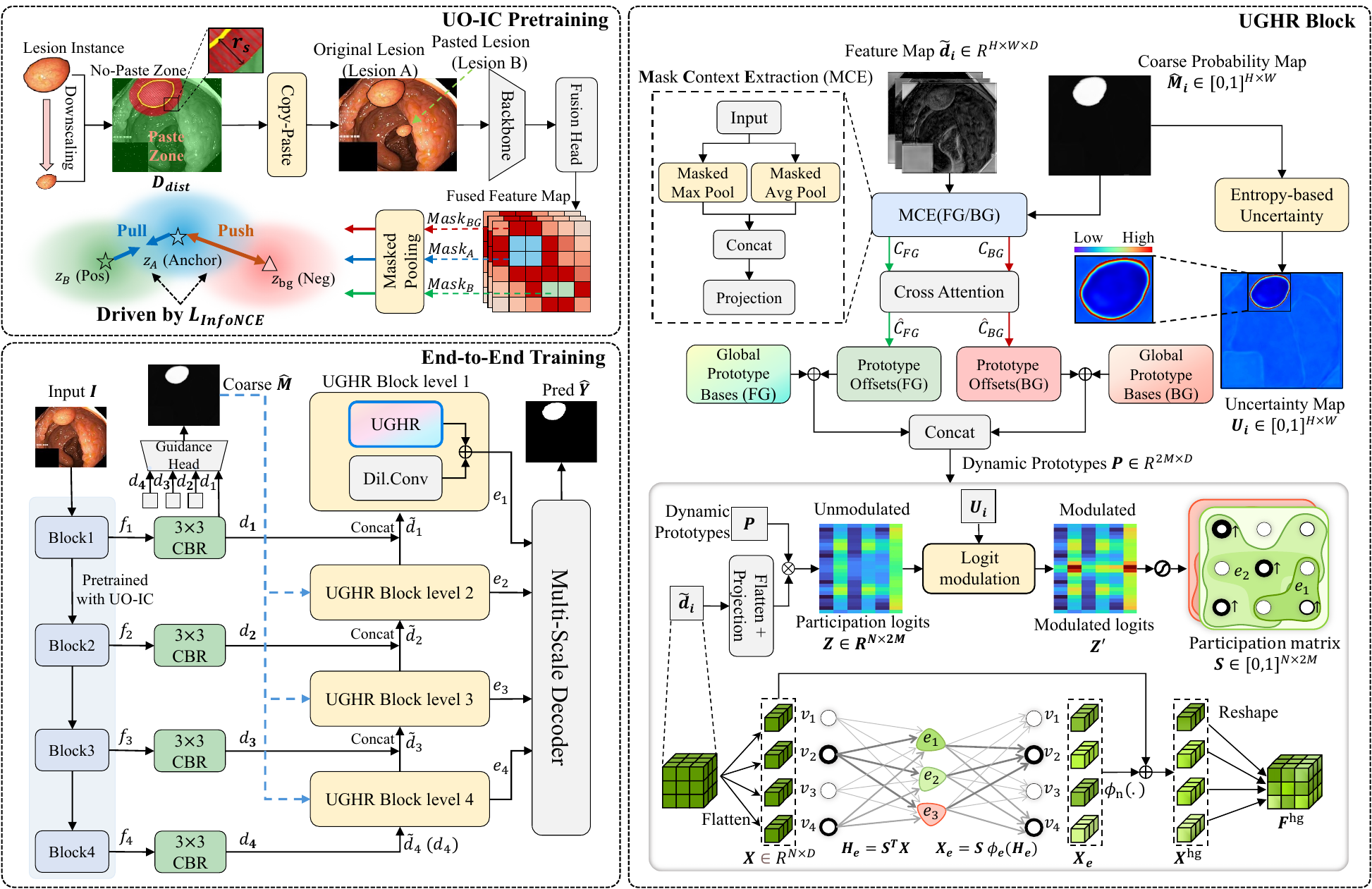}
  \caption{Overall framework of the proposed UHR-Net. The upper left part is the UO-IC pretraining stage. The lower left part is the end-to-end training stage, encompassing the complete network. The right part is the detailed structure of the UGHR block.}
  \label{fig:framework}
\end{figure}

\section{Methodology}
\label{sec:method}

\subsection{Overview}
UHR-Net couples instance-level discriminative pretraining with uncertainty-guided structured refinement to alleviate small-lesion cue dilution, lesion-like background interference, and unstable predictions in ambiguous regions.

As illustrated in Fig.~\ref{fig:framework}, the network predicts a foreground probability map $\hat{Y}$ from the input image. Training is organized into UO-IC pretraining followed by end-to-end optimization. In the pretraining phase, copy-paste augmentation forms lesion pairs and lesion-like background samples are used as hard negatives for contrastive learning. In the segmentation phase, a guidance head generates a coarse probability map $\hat{M}$, from which an entropy-based uncertainty map $U$ is computed. This uncertainty cue is passed to multi-scale UGHR blocks, where prototype-based hypergraph reasoning and a local convolutional branch jointly refine decoder features before producing the final prediction $\hat{Y}$. Details are provided in Section~\ref{sec:ughr}.

\subsection{Uncertainty-Oriented Instance Contrastive pretraining}
\label{sec:uoic}
To alleviate small-lesion cue dilution and lesion-like background interference, we propose UO-IC, an instance-level contrastive pretraining strategy that combines geometrically constrained copy-paste augmentation with lesion-like hard-negative mining.

Given the ground-truth mask $Y$, we randomly select a lesion instance as Lesion A and generate a scaled replica Lesion B. To avoid overlap with existing foreground regions, we compute a Euclidean distance map $D_{dist}$ on the background region $1-Y$ and select a paste center $p$ satisfying $D_{dist}(p) \geq r_s$, where $r_s$ is the bounding-box circumradius of Lesion B. We paste Lesion B at this location to obtain the augmented image $I^{cp}$ and supervision mask $Y^{cp}$, while retaining the single-instance masks $M_A$ and $M_B$. Given the feature map $\tilde{F}$ extracted from $I^{cp}$, masked average pooling over $M_A$ and $M_B$ yields $\mathbf{z}_A$ and $\mathbf{z}_B$, which form the positive pair for InfoNCE.

To construct hard negatives, we focus on lesion-like background locations in the background region $1-Y^{cp}$ where the predicted foreground probability map $\hat{Y}^{cp}$ is high. Specifically, we perform weighted masked average pooling (WMAP) on features using $\hat{Y}^{cp}$ as the weighting term, obtaining the lesion-like background representation
\begin{equation}
\mathbf{z}_{bg}=\mathrm{WMAP}\!\left(\tilde{F},\; (1-Y^{cp})\odot \hat{Y}^{cp}\right).
\label{eq:zbg}
\end{equation}
During training, we compute one lesion-like background representation per sample and use the in-batch set $\mathbf{z}_{bg}^{(k)}$ as hard negatives to contrast against $\mathbf{z}_A$.

Based on the above representations, we adopt the InfoNCE loss for instance-level contrastive optimization. We take $\mathbf{z}_A$ as the anchor, $\mathbf{z}_B$ as the positive sample, and $\{\mathbf{z}_{bg}^{(k)}\}_{k=1}^{N_{neg}}$ as negatives, where $N_{neg}$ denotes the number of in-batch negatives. Let $\mathrm{sim}(\cdot,\cdot)$ denote cosine similarity, and the InfoNCE objective is defined as
\begin{equation}
\label{eq:infonce_logits}
\begin{split}
s^{+} &= \mathrm{sim}(\mathbf{z}_A,\mathbf{z}_B)/\tau,\\
s^{-}_k &= \mathrm{sim}(\mathbf{z}_A,\mathbf{z}_{bg}^{(k)})/\tau.
\end{split}
\end{equation}
\begin{equation}
\label{eq:infonce}
\mathcal{L}_{\mathrm{InfoNCE}}
= -\log\left[
\frac{\exp(s^{+})}{\exp(s^{+})+\sum_{k=1}^{N_{neg}}\exp(s^{-}_k)}
\right].
\end{equation}
This objective pulls $\mathbf{z}_A$ closer to its scaled-and-pasted replica $\mathbf{z}_B$ while pushing $\mathbf{z}_A$ away from the lesion-like background representations. Here, $\tau$ is the temperature parameter that scales the contrastive logits.

Moreover, to avoid degrading pixel-wise localization by relying solely on instance-level contrastive constraints, we also impose segmentation supervision during pretraining; the complete objective is provided in Section~\ref{sec:loss}.

\subsection{Uncertainty-Guided Hypergraph Refinement block}
\label{sec:ughr}
Along the refinement path, the UGHR block at the $i$-th scale takes a fused input feature $\tilde{d}_i$, which is obtained by combining the channel-aligned feature at the current scale with cross-scale information from the deeper refined output. Meanwhile, the coarse segmentation probability map $\hat{M}$ is resized to this scale to obtain $\hat{M}_i$. Given $\tilde{d}_i$ and $\hat{M}_i$, the UGHR block outputs the refined feature $e_i$. Since ambiguous regions typically have higher uncertainty and larger prediction fluctuations, the UGHR branch derives a pixel-wise uncertainty map $U_i$ from $\hat{M}_i$ and uses it to modulate the node-hyperedge participation logits before normalization. This modulation amplifies the participation logits at highly uncertain locations, encouraging larger normalized participation weights after normalization, so that subsequent hypergraph message passing can more effectively aggregate contextual information from these regions, improving prediction consistency in ambiguous regions. Next, we introduce uncertainty-guided hypergraph construction and hypergraph message passing and feature updating in turn.

\subsubsection{Uncertainty-Guided Hypergraph Construction.}
We compute a pixel-wise uncertainty map $U_i$ from the coarse segmentation probability map $\hat{M}_i$:
\begin{equation}
U_i
=
-\frac{\hat{M}_i\log(\hat{M}_i+\varepsilon)+(1-\hat{M}_i)\log(1-\hat{M}_i+\varepsilon)}{\log 2},
\label{eq:uncertainty}
\end{equation}
where $\varepsilon$ is a numerical stability constant, and $\log 2$ normalizes the binary entropy to $[0,1]$. When $\hat{M}_i$ is close to 0.5, $U_i$ attains larger values, which typically correspond to ambiguous regions. In what follows, $U_i$ is used to modulate the node-hyperedge participation logits, yielding normalized participation weights.

Next, we treat each pixel location in $\tilde{d}_i$ as a node and introduce two groups of hyperedge prototypes, i.e., foreground and background, so that hyperedges can selectively aggregate node information over the entire image. Prototype generation is conditioned on the coarse segmentation prior: we use $\hat{M}_i$ and $1-\hat{M}_i$ as soft weights for the foreground and background, respectively, and adopt a lightweight Mask Context Extraction (MCE) module to perform mask-guided pooling on $\tilde{d}_i$, producing a foreground context vector $c_{FG}$ and a background context vector $c_{BG}$. Considering that coarse predictions in ambiguous regions can be mixed, we further apply bidirectional cross-attention between the two contexts to obtain $\tilde{c}_{FG}$ and $\tilde{c}_{BG}$, which are used to generate sample-wise offsets and thus construct sample-wise dynamic prototypes.

Specifically, we maintain two groups of learnable global prototype bases $P^g_{FG}$ and $P^g_{BG}$, where each group contains $M$ hyperedge prototypes and $P^g_{FG}, P^g_{BG}\in\mathbb{R}^{M\times D}$. Here, $D$ denotes the node embedding dimension, which is the channel dimension of $\tilde{d}_i$. Given the interacted contexts $\tilde{c}_{FG}$ and $\tilde{c}_{BG}$, we generate sample-wise prototype offsets $\Delta P_{FG}, \Delta P_{BG}\in\mathbb{R}^{M\times D}$, and add them to the corresponding bases to obtain dynamic foreground and background prototypes:
\begin{equation}
P_{FG}=P^{g}_{FG}+\Delta P_{FG},\qquad
P_{BG}=P^{g}_{BG}+\Delta P_{BG}.
\label{eq:dynamic_proto}
\end{equation}
We then concatenate the two prototype sets to obtain the sample-specific dynamic prototypes
\begin{equation}
P=\mathrm{Concat}(P_{FG},P_{BG})=[p_1,\ldots,p_{2M}]\in\mathbb{R}^{2M\times D}.
\label{eq:concat_proto}
\end{equation}
where the first $M$ prototypes belong to the foreground group and the remaining $M$ prototypes belong to the background group. The dynamic prototypes $P$ are used to compute node-hyperedge participation logits.

To explicitly inject uncertainty into node-hyperedge participation assignment, let $u_n\in[0,1]$ denote the $n$-th element of the flattened uncertainty map $U_i$. After obtaining $P$, we flatten $\tilde{d}_i$ along the spatial dimensions into a node feature matrix $X=[x_1,\ldots,x_N]^T\in\mathbb{R}^{N\times D}$, and compute unnormalized node-hyperedge logits via scaled dot-product:
\begin{equation}
z_{n,m}=\frac{x_n^T p_m}{\sqrt{D}},\qquad n=1,\ldots,N,\; m=1,\ldots,2M.
\label{eq:assoc_logit}
\end{equation}
We then modulate $z_{n,m}$ pointwise by $u_n$ and normalize over the node dimension for each hyperedge to obtain the participation matrix $S\in\mathbb{R}^{N\times 2M}$:
\begin{equation}
\begin{aligned}
z'_{n,m} &= z_{n,m}\cdot 2^{\beta u_n},\\
S_{n,m} &= \frac{\exp(z'_{n,m})}{\sum_{n'=1}^{N}\exp(z'_{n',m})},\qquad m=1,\ldots,2M,
\end{aligned}
\label{eq:participation}
\end{equation}
where $S_{n,m}$ denotes the normalized participation weight of node $n$ in hyperedge $m$, and $\beta$ is the modulation strength. This uncertainty-aware logit modulation increases the normalized participation weights of high-uncertainty nodes in $S$, thereby emphasizing their contributions during subsequent hypergraph message passing and feature updates.

\subsubsection{Hypergraph Message Passing and Feature Updating.}
After obtaining the participation matrix $S$ and the node features $X$, we perform a two-step message passing scheme on the hypergraph to update node representations. First, nodes are aggregated into hyperedge representations using the participation matrix $S$, $H_e=S^T X$. Then, $H_e$ is projected by an edge mapping $\phi_e(\cdot)$ and propagated back to nodes using $S$, yielding the returned messages $X_e$. The messages are further transformed by a node mapping $\phi_n(\cdot)$ and added to the input node features $X$ via a residual connection to obtain the updated node features $X^{hg}$:
\begin{equation}
X_e = S\phi_e(H_e),\qquad X^{hg}=X+\phi_n(X_e).
\label{eq:message_passing}
\end{equation}
Reshaping $X^{hg}$ back to the spatial layout gives the output of the UGHR branch, denoted as $F^{hg}$. Meanwhile, the Dil.Conv branch produces a locally enhanced feature map $F^{conv}$. We fuse the two branch outputs in a pixel-wise manner to obtain the refined feature $e_i$.

\subsection{Loss Function}
\label{sec:loss}
Training consists of two phases: UO-IC pretraining and end-to-end training. In both phases, we use a combined Dice and BCE loss for segmentation supervision, denoted as $\mathcal{L}_{seg}(\hat{Y},Y)$.

During pretraining, we incorporate the instance-level contrastive term in Section~\ref{sec:uoic} on top of the segmentation supervision. The pretraining objective is
\begin{equation}
\mathcal{L}_{pre}
=
\mathcal{L}_{seg}(\hat{Y}^{cp},Y^{cp})
+
\lambda_{ic}\mathcal{L}_{\mathrm{InfoNCE}}.
\label{eq:loss_pre}
\end{equation}
During end-to-end training, in addition to producing the final prediction $\hat{Y}$, the guidance head also outputs a coarse segmentation probability map $\hat{M}$. To keep the coarse guidance stable during training and encourage it to learn a coarse prior aligned with the final segmentation objective, we apply an auxiliary segmentation supervision to the upsampled coarse map $\hat{M}^{\uparrow}$ (upsampled to the output resolution). The overall training objective is
\begin{equation}
\mathcal{L}_{train}
=
\mathcal{L}_{seg}(\hat{Y},Y)
+
\lambda_{aux}\mathcal{L}_{seg}(\hat{M}^{\uparrow},Y).
\label{eq:loss_all}
\end{equation}
where $\lambda_{ic}$ and $\lambda_{aux}$ denote the weights of the contrastive term and the auxiliary supervision, respectively.

\section{Experiments}
\label{sec:exp}

\begin{table}[H]
\centering
\caption{Comparison with other methods on the Kvasir-Sessile, Kvasir-SEG and GlaS datasets. Bold indicates the best result and Underlined indicates the second best.}
\label{tab:three_datasets}
\small
\begingroup
\setlength{\tabcolsep}{2.0pt}
\renewcommand{\arraystretch}{0.95}
\resizebox{\textwidth}{!}{%
\begin{tabular}{lcccccccccccc}
\toprule
\multirow{2}{*}{Methods} &
\multicolumn{4}{c}{Kvasir-Sessile} &
\multicolumn{4}{c}{Kvasir-SEG} &
\multicolumn{4}{c}{GlaS} \\
\cmidrule(lr){2-5}\cmidrule(lr){6-9}\cmidrule(lr){10-13}
 & mIoU & mDSC & Rec. & Prec. &
   mIoU & mDSC & Rec. & Prec. &
   mIoU & mDSC & Rec. & Prec. \\
\midrule
U-Net~\cite{ronneberger2015u}      & 23.1 & 33.8 & 45.1 & 46.6 & 65.5 & 75.8 & 83.6 & 77.6 & 75.8 & 85.5 & 90.3 & 82.8 \\
U-Net++~\cite{zhou2018unetpp}      & 38.4 & 50.2 & 62.5 & 51.8 & 67.9 & 77.2 & 86.5 & 77.6 & 77.6 & 86.9 & 89.6 & 85.5 \\
PraNet~\cite{fan2020pranet}        & 66.7 & 77.4 & 80.7 & 82.4 & 83.0 & 89.4 & 90.6 & 91.3 & 71.8 & 83.0 & 90.9 & 78.0 \\
TGANet~\cite{tomar2022tganet}      & 74.4 & 82.0 & 79.3 & 85.9 & 83.3 & 89.8 & 91.3 & 91.2 & 71.8 & 84.7 & 86.9 & 80.2 \\
DCSAU-Net~\cite{xu2023dcsau}       & 72.6 & 81.1 & 65.6 & 62.9 & 83.5 & 88.9 & 89.5 & 89.5 & 77.6 & 86.5 & 93.0 & 82.5 \\
CASF-Net~\cite{zheng2023casf}      & 60.5 & 72.4 & 78.0 & 74.8 & 81.7 & 88.7 & 89.2 & 88.2 & 78.4 & 87.2 & 91.3 & 85.9 \\
DTAN~\cite{zhao2024dtan}           & 76.4 & 84.2 & 84.2 & 85.9 & 84.1 & 90.4 & 91.6 & \best{92.0} & 78.5 & 87.9 & 85.8 & 90.2 \\
CMUNeXt~\cite{tang2024cmunext}     & 75.3 & 85.2 & 89.6 & 82.9 & 81.1 & 89.0 & 89.3 & 89.9 & 81.2 & 89.1 & 89.2 & 90.1 \\
ESPNet~\cite{toman2025espnet}
           & \second{82.4} & \second{90.4} & 89.3 & \best{91.4}
           & 83.6 & \second{90.8} & 90.9 & 91.2
           & 85.0 & \second{91.9} & \second{94.3} & 89.6 \\
ConDSeg~\cite{lei2025condseg}
           & 81.2 & 89.1 & \second{90.1} & \second{90.0}
           & \second{84.6} & 90.5 & \second{92.3} & \second{91.7}
           & \second{85.1} & 91.6 & 93.5 & \second{90.5} \\
\midrule
\textbf{Ours}
           & \best{83.5} & \best{90.6} & \best{91.4} & \best{91.4}
           & \best{85.0} & \best{90.8} & \best{92.7} & 91.2
           & \best{87.0} & \best{92.7} & \best{94.7} & \best{91.5} \\
\bottomrule
\end{tabular}%
}
\endgroup
\end{table}

\begin{table}[H]
\centering
\caption{Comparison with other methods on the ISIC-2016 and ISIC-2017 datasets. Same notation as Table~\ref{tab:three_datasets}.}
\label{tab:two_datasets}
\small
\begingroup
\setlength{\tabcolsep}{3pt}
\renewcommand{\arraystretch}{0.95}
\begin{tabular*}{\textwidth}{@{\extracolsep{\fill}}lcccc@{}}
\toprule
\multirow{2}{*}{Methods} &
\multicolumn{2}{c}{ISIC-2016} &
\multicolumn{2}{c}{ISIC-2017} \\
\cmidrule(lr){2-3}\cmidrule(lr){4-5}
 & mIoU & mDSC & mIoU & mDSC \\
\midrule
U-Net~\cite{ronneberger2015u}     & 83.6 & 90.3 & 73.7 & 82.8 \\
CENet~\cite{gu2019net}     & 84.6 & 90.9 & 76.4 & 84.8 \\
FAT-Net~\cite{wu2022fat}   & 85.3 & 91.6 & 76.5 & 85.0 \\
DCSAU-Net~\cite{xu2023dcsau} & 85.3 & 91.4 & 76.1 & 85.0 \\
BGDiffSeg~\cite{guo2024bgdiffseg} & 85.5 & 92.2 & 79.7 & 88.7 \\
MSCB-UNet~\cite{wang2025mscb} & 84.2 & 91.4 & \second{81.2} & \best{89.6} \\
ConDSeg~\cite{lei2025condseg}   & \second{86.8} & \second{92.5} & 80.9 & 88.3 \\
\midrule
\textbf{Ours}
          & \best{87.6} & \best{92.9} & \best{81.8} & \second{89.2} \\
\bottomrule
\end{tabular*}
\endgroup
\end{table}

\subsection{Experimental Setup}
\label{sec:exp_setup}
We evaluate UHR-Net on five public datasets: Kvasir-SEG~\cite{jha2020kvasir}, Kvasir-Sessile~\cite{jha2021comprehensive}, GlaS~\cite{sirinukunwattana2017gland}, ISIC-2016~\cite{gutman2016skin}, and ISIC-2017~\cite{codella2018skin} following the same split and setting as~\cite{lei2025condseg}. All experiments are implemented using PyTorch. For the UO-IC pretraining stage, we set $\tau=0.10$ and $\lambda_{ic}=1.0$. For the end-to-end training stage, we set $\lambda_{aux}=0.1$. In UGHR block, we set $M=8$ prototypes per group and $\beta=1.0$. For both stages, we set the batch size to 16 and use the Adam optimizer with a learning rate of $1\times10^{-4}$.

\subsection{Comparison with Other Methods}
\label{sec:exp_compare}
\begin{figure}[H]
    \centering
    \includegraphics[width=0.86\textwidth]{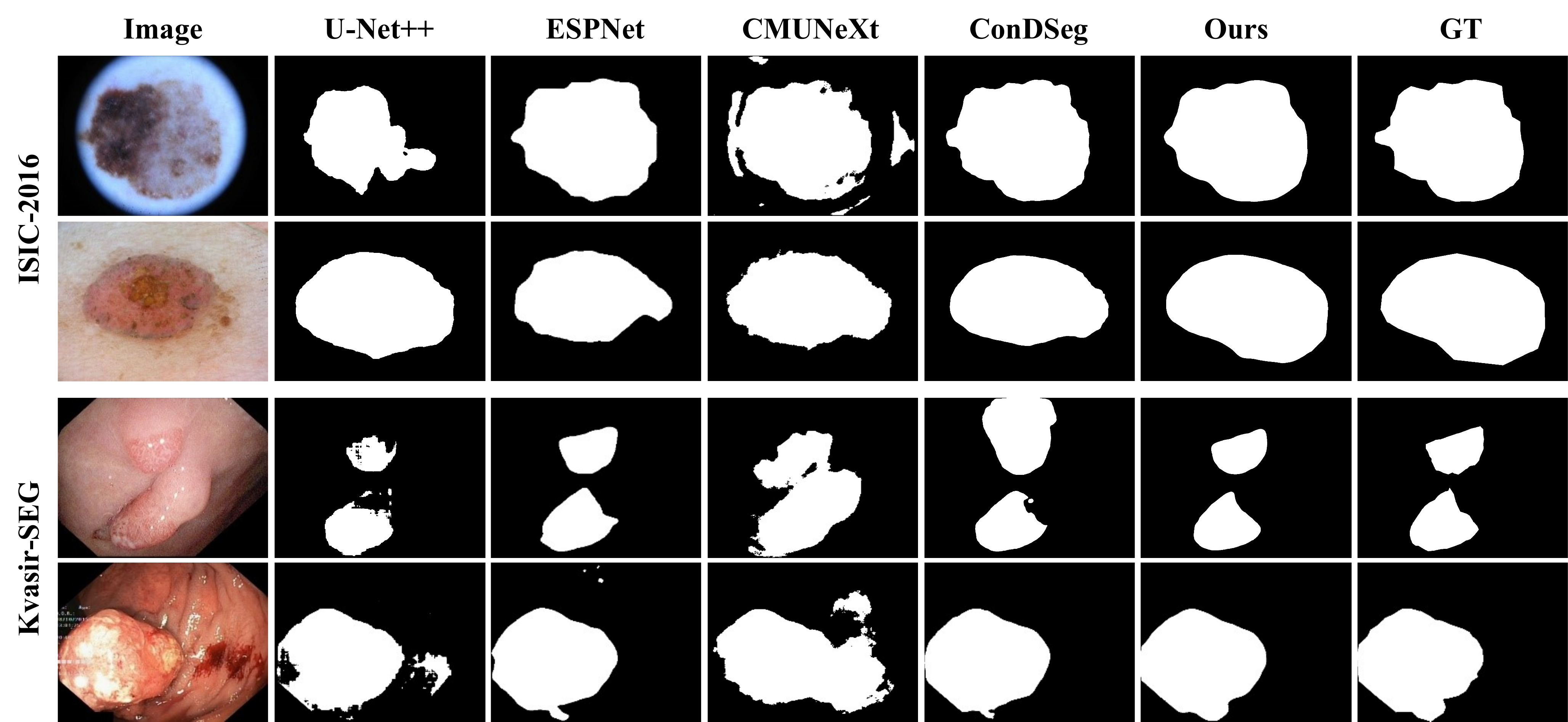}
    \caption{Qualitative comparisons on ISIC-2016 and Kvasir-SEG. Columns show Image, U-Net++, ESPNet, CMUNeXt, ConDSeg, Ours, and Ground Truth (GT).}
    \label{fig:qual_comp}
\end{figure}

Tables~\ref{tab:three_datasets} and~\ref{tab:two_datasets} summarize the quantitative results on the five benchmarks. UHR-Net obtains the highest mIoU on all datasets and maintains strong mDSC performance, covering endoscopic, glandular, and dermoscopic segmentation scenarios. These results suggest that the proposed pretraining and refinement designs improve both lesion discrimination and boundary-level correction. The visual examples in Fig.~\ref{fig:qual_comp} also show that our method better preserves small lesion regions and suppresses confusing background responses.

\subsection{Ablation Study}
\label{sec:exp_ablation}
All ablations are conducted on Kvasir-SEG. Table~\ref{tab:ablation_hyperparams} analyzes the sensitivity of $M$, $\beta$, $\tau$, and $\lambda_{ic}$, and Table~\ref{tab:ablation_components} evaluates the proposed components.

\subsubsection{Effectiveness of proposed components.}
In Table~\ref{tab:ablation_components}, Exp.~1 is the baseline, Exp.~2 adds UO-IC, Exps.~3--6 evaluate the UGHR branch and its two key designs, and Exp.~7 is the full model. When Base HR is disabled, the UGHR branch is replaced by a standard convolution; when Unc. Guid. is disabled, $U$ no longer modulates participation logits; when FG/BG HG is disabled, a shared prototype set is used.
\enlargethispage{5\baselineskip}

\begin{table}[H]
\centering
\caption{Hyperparameter sensitivity analysis on the Kvasir-SEG dataset. Only mIoU and mDSC are reported for compactness. The best results in each group are shown in bold.}
\label{tab:ablation_hyperparams}
\small
\begingroup
\setlength{\tabcolsep}{2.2pt}
\renewcommand{\arraystretch}{0.95}
\resizebox{\textwidth}{!}{%
\begin{tabular}{lcc@{\hspace{10pt}}lcc@{\hspace{10pt}}lcc@{\hspace{10pt}}lcc}
\toprule
\multicolumn{3}{c}{Prototype number $M$} & \multicolumn{3}{c}{Modulation strength $\beta$} & \multicolumn{3}{c}{Temperature $\tau$} & \multicolumn{3}{c}{Contrastive weight $\lambda_{ic}$} \\
\cmidrule(lr){1-3}\cmidrule(lr){4-6}\cmidrule(lr){7-9}\cmidrule(lr){10-12}
Value & mIoU & mDSC & Value & mIoU & mDSC & Value & mIoU & mDSC & Value & mIoU & mDSC \\
\midrule
2  & 83.5 & 89.5 & 0.1 & 83.7 & 89.9 & 0.01 & 79.8 & 87.0 & 0.1 & 83.5 & 89.6 \\
4  & 84.0 & 90.3 & 0.5 & 84.2 & 90.1 & 0.05 & 83.7 & 89.4 & 0.5 & 83.8 & 89.7 \\
\textbf{8}  & \textbf{85.0} & \textbf{90.8} & \textbf{1.0} & \textbf{85.0} & \textbf{90.8} & \textbf{0.10} & \textbf{85.0} & \textbf{90.8} & \textbf{1.0} & \textbf{85.0} & \textbf{90.8} \\
16 & 84.7 & 89.9 & 1.5 & 84.3 & 89.7 & 0.20 & 82.5 & 89.2 & 2.0 & 84.2 & 90.1 \\
32 & 84.5 & 89.9 & 2.0 & 84.1 & 89.9 & 0.50 & 81.5 & 88.5 & 5.0 & 82.5 & 88.8 \\
\bottomrule
\end{tabular}%
}
\endgroup
\end{table}

\begin{table}[H]
\centering
\caption{Ablation study on key components. Checkmarks indicate enabled components. Abbreviations: ``Base HR'' stands for Base Hypergraph Refiner; ``Unc. Guid.'' stands for Uncertainty Guidance; ``FG/BG HG'' stands for FG/BG Hyperedge Groups.}
\label{tab:ablation_components}
\setlength{\tabcolsep}{2.2pt}
\renewcommand{\arraystretch}{0.95}
\small
\resizebox{\textwidth}{!}{%
\begin{tabular}{c c c c c c c}
\toprule
 & \multicolumn{1}{c}{Pretraining} & \multicolumn{3}{c}{UGHR Branch} & \multicolumn{2}{c}{Metric} \\
\cmidrule(lr){2-2} \cmidrule(lr){3-5} \cmidrule(lr){6-7}
 & UO-IC
& \multicolumn{1}{c}{\shortstack{Base HR}}
& \multicolumn{1}{c}{\shortstack{Unc.\ Guid.}}
& \multicolumn{1}{c}{\shortstack{FG/BG HG}}
 & mIoU & mDSC \\
\midrule
(1) & \xmark & \xmark & \xmark & \xmark & 81.2 & 88.8 \\
(2) & \cmark & \xmark & \xmark & \xmark & 82.2 & 89.0 \\
(3) & \xmark & \cmark & \xmark & \xmark & 83.0 & 89.5 \\
(4) & \xmark & \cmark & \xmark & \cmark & 83.5 & 89.8 \\
(5) & \xmark & \cmark & \cmark & \xmark & 83.9 & 90.1 \\
(6) & \xmark & \cmark & \cmark & \cmark & 84.4 & 90.3 \\
(7) & \cmark & \cmark & \cmark & \cmark & \best{85.0} & \best{90.8} \\
\bottomrule
\end{tabular}%
}
\end{table}

The results show that UO-IC improves the baseline by enhancing instance-level discrimination, while Base HR validates the benefit of higher-order aggregation. Unc. Guid. and FG/BG HG bring further gains by focusing message passing on ambiguous regions and reducing foreground-background interference, respectively. Their combination with UO-IC yields the best performance, indicating that UO-IC and UGHR are complementary.

\subsubsection{Hyperparameter sensitivity.}
Table~\ref{tab:ablation_hyperparams} summarizes the sensitivity analysis of four key hyperparameters on Kvasir-SEG. For the hyperedge prototype number, increasing $M$ from 2 to 8 improves performance, while further increasing it brings slight degradation, suggesting that excessive prototypes may introduce redundant hyperedges. For uncertainty modulation, $\beta=1.0$ achieves the best result, indicating a balanced strength for emphasizing uncertain nodes. For UO-IC pretraining, $\tau=0.10$ and $\lambda_{ic}=1.0$ yield the best performance, showing that moderate contrastive sharpness and loss weighting are sufficient for stable representation learning. Therefore, we set $M=8$, $\beta=1.0$, $\tau=0.10$, and $\lambda_{ic}=1.0$ as the default configuration.

\subsection{Qualitative Validation of Proposed Designs}
\label{sec:exp_visualization}
We qualitatively validate UO-IC and UGHR through t-SNE representations and entropy-based uncertainty maps.
\enlargethispage{12\baselineskip}

\begin{figure}[H]
    \centering
    \includegraphics[width=0.60\textwidth]{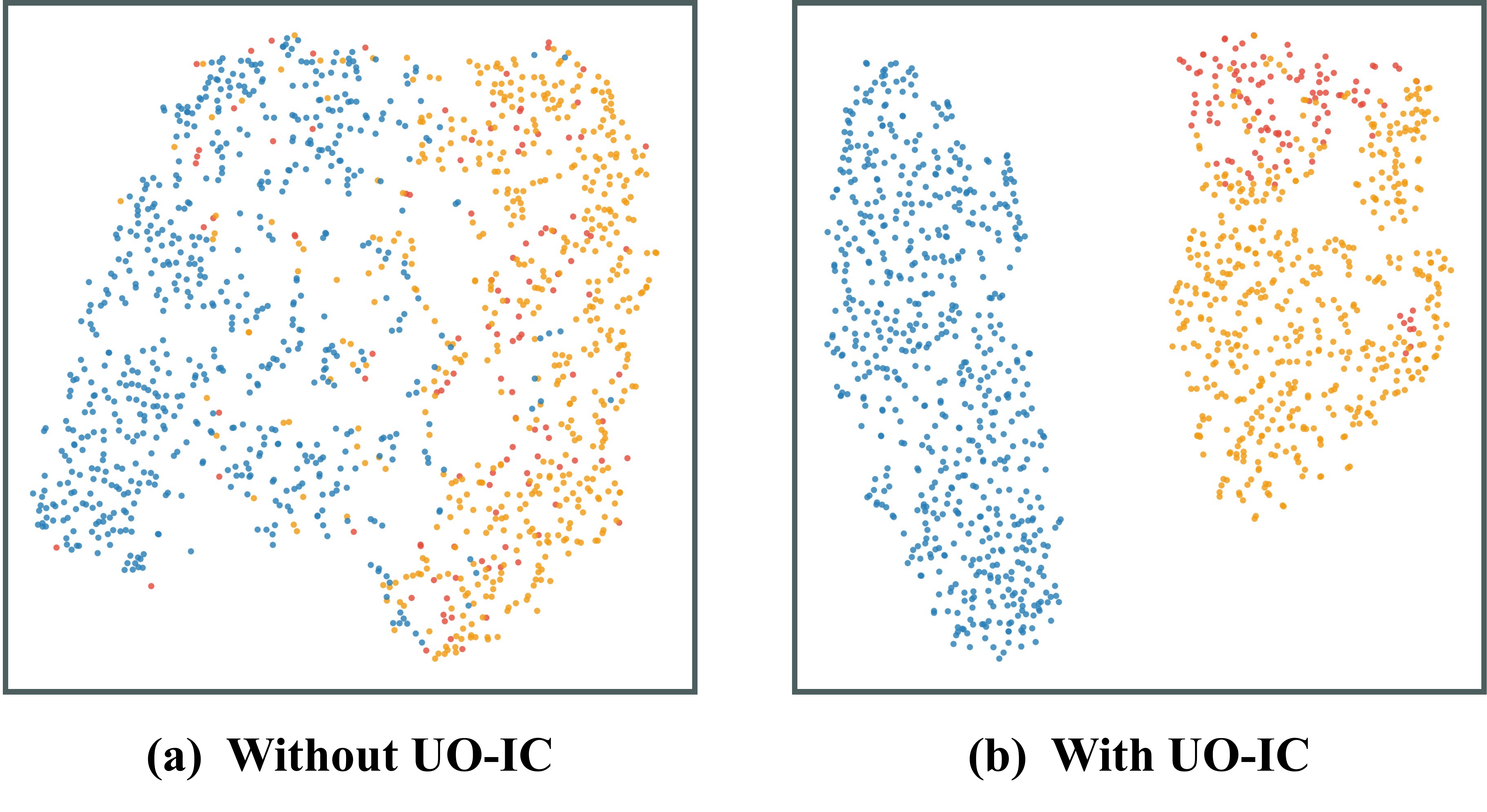}
    \caption{t-SNE visualization of representations learned with and without UO-IC. Blue: lesion-like background hard negatives; red: small-lesion representations; yellow: remaining lesion representations.}
    \label{fig:tsne_compare}
\end{figure}

\begin{figure}[H]
    \centering
    \includegraphics[width=0.95\textwidth]{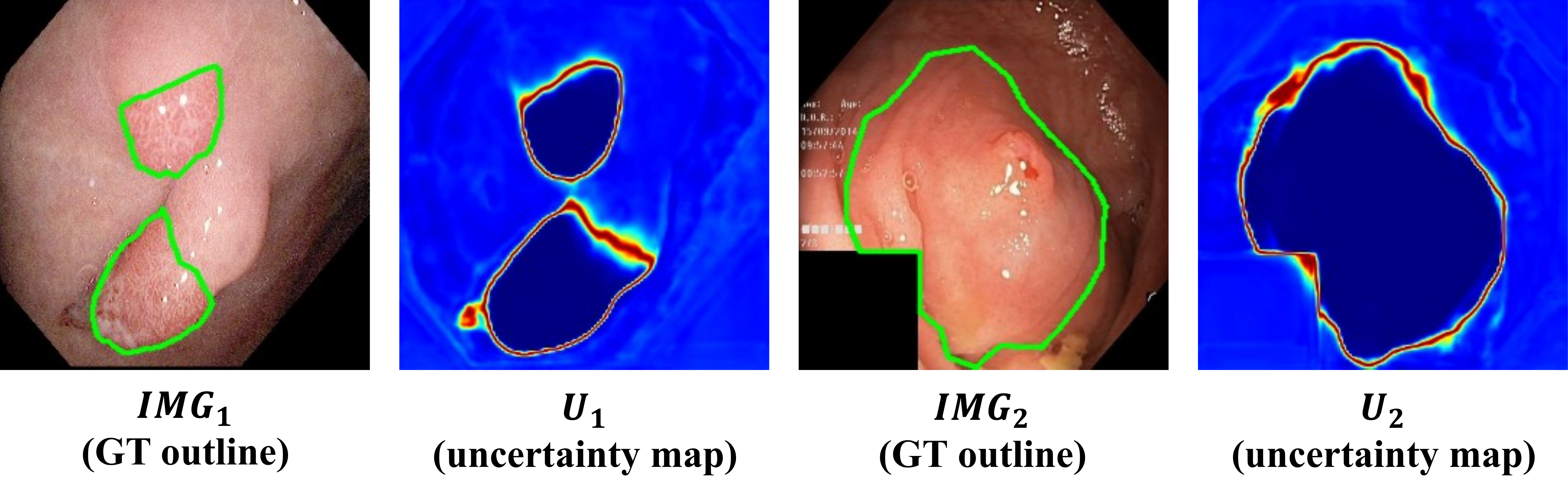}
    \caption{Entropy-based uncertainty maps derived from coarse probability maps. Brighter regions indicate higher uncertainty.}
    \label{fig:uncertainty_vis}
\end{figure}

As shown in Fig.~\ref{fig:tsne_compare}, red points denote small-lesion representations, defined as original-lesion representations whose resized lesion areas fall in the bottom 20\%, while yellow points denote the remaining lesion representations. UO-IC makes small-lesion representations more compact and better aligned with other lesion representations, while separating lesion-like background hard negatives from lesion clusters. This supports its role in alleviating small-lesion cue dilution and background interference.

Fig.~\ref{fig:uncertainty_vis} shows that high uncertainty mainly appears around lesion boundaries and transition regions, supporting the use of entropy-derived uncertainty to guide UGHR toward ambiguous regions where stronger node-hyperedge participation and higher-order aggregation are needed.

\FloatBarrier
\section{Conclusion}
In this paper, we present UHR-Net, an Uncertainty-Aware Hypergraph Refinement Network for medical lesion segmentation that tackles cue dilution in small lesions and prediction instability in ambiguous regions. We introduce UO-IC pretraining to enhance instance discriminability via geometry-aware copy-paste and lesion-like hard-negative mining. We further develop UGHR blocks that (i) derive an entropy-based uncertainty map to modulate node-hyperedge participation logits, and (ii) employ foreground- and background-conditioned hyperedge prototypes to decouple higher-order interactions. Extensive experiments on five public benchmarks demonstrate consistent gains over representative methods.

\begin{credits}
\subsubsection{Acknowledgments.}
This work was supported by the National Natural Science Foundation of China (No. 62176242), Hubei Provincial Natural Science Foundation of China (No. 2025AFB832) and Open Research Fund of State Key Laboratory of Information Engineering in Surveying, Mapping and Remote Sensing, Wuhan University (No. 23E03), also in part by the Opening Fund of Key Laboratory of Geological Survey and Evaluation of Ministry of Education (No. GLAB2024ZR09) and the Fundamental Research Funds for the Central Universities.

\subsubsection{Disclosure of Interests.}
The authors have no competing interests to declare that are relevant to the content of this article.
\end{credits}

\end{document}